\documentclass[11pt]{article}

\usepackage[utf8]{inputenc}
\usepackage[T1]{fontenc}
\usepackage{times}
\usepackage{graphicx}
\usepackage{booktabs}
\usepackage{hyperref}
\usepackage{hyperxmp} 
\usepackage{url}
\usepackage{xurl} 
\usepackage{amsmath}
\usepackage{multirow}
\usepackage[margin=1in]{geometry}

\hypersetup{
  colorlinks=true,
  linkcolor=blue,
  citecolor=blue,
  urlcolor=blue,
  pdftitle={WCXB: A Multi-Type Web Content Extraction Benchmark},
  pdfauthor={Murrough Foley},
  pdfsubject={A 2008-page benchmark of web content extraction across 7 structural page types},
  pdfkeywords={web content extraction, benchmark, evaluation, information retrieval, natural language processing, dataset, HTML parsing, boilerplate removal, web mining, main content detection},
  pdfcopyright={Copyright (c) 2026 Murrough Foley. Licensed under CC-BY-4.0.},
  pdflicenseurl={https://creativecommons.org/licenses/by/4.0/}
}

\title{WCXB: A Multi-Type Web Content Extraction Benchmark}
\author{Murrough Foley \\
  ORCID: \href{https://orcid.org/0009-0008-3127-2101}{0009-0008-3127-2101} \\
  \texttt{murrough@murroughfoley.com} \\
  \url{https://murroughfoley.com}
}
\date{}

\begin{document}
\maketitle

\begin{abstract}
Web content extraction --- isolating a page's main content from surrounding boilerplate --- is a prerequisite for search indexing, retrieval-augmented generation, NLP dataset construction, and large language model training. Progress in this area has been constrained by the limitations of existing evaluation benchmarks, which are small (100--800 pages), restricted to news articles, or based on web pages from over a decade ago. We introduce the Web Content Extraction Benchmark (WCXB), a dataset of 2,008 web pages from 1,613 domains spanning seven structurally distinct page types: articles, forums, products, collections, listings, documentation, and service pages. The dataset includes a 1,497-page development set and a 511-page held-out test set with matched page type distributions. Ground truth annotations were produced through a five-stage pipeline: LLM-assisted drafting, automated verification, four-pass frontier model review, snippet and quality verification scripts, and human review. We evaluate 13 extraction systems --- 11 heuristic and 2 neural --- and find that while top systems converge on articles (F1~=~0.93), performance diverges sharply on structured page types (F1~=~0.41--0.84), revealing blind spots invisible to existing article-only benchmarks. The dataset is released under CC-BY-4.0 with HTML source files, ground truth annotations, page type labels, and baseline results.
\end{abstract}

\section{Introduction}

Content extraction from web pages is foundational infrastructure for the modern web. Search engines rely on extracted content for indexing and ranking. Retrieval-augmented generation (RAG) systems require clean text from web sources. NLP researchers construct training corpora by extracting content at scale. And increasingly, large language model training pipelines process billions of web pages, where extraction quality directly impacts downstream model performance.

The dominant approach treats extraction as a single task: given an HTML document, identify and return the main content while discarding navigation menus, advertisements, sidebars, cookie banners, footers, and other boilerplate elements. Over two decades of research has produced systems such as Boilerpipe~\cite{boilerpipe}, Readability~\cite{readability}, jusText~\cite{justext}, and Trafilatura~\cite{trafilatura}, with recent neural approaches including MinerU-HTML~\cite{mineruhtml} and ReaderLM-v2~\cite{readerlm}.

Progress in content extraction has been measured against a small number of benchmarks: CleanEval (797 pages, 2007)~\cite{cleaneval}, Dragnet (143 articles, 2013)~\cite{dragnet}, the ScrapingHub benchmark (181 articles, 2019)~\cite{scrapinghub}, and Google-Trends-2017 (180 pages)~\cite{boilernet}. Bevendorff et al.~\cite{bevendorff} combined eight such datasets into approximately 3,985 pages and evaluated 14 extractors, finding that ``performance is quite genre-dependent and no single extractor performs best on all types of web pages.'' Yet their combined dataset inherits the limitations of its sources --- primarily news articles, with dated HTML from an earlier web era --- and they did not create new annotations to address the gap.

This genre-dependence observation is critical but undertested. The modern web is not composed solely of articles. A web crawler encounters product pages with structured specifications, discussion forums with nested user posts, documentation pages with code blocks and sidebar navigation, marketing pages with content distributed across hero sections and testimonials, and category pages with product grids. Each page type has fundamentally different HTML structure, and heuristics optimized for article extraction fail on them in predictable ways. No existing benchmark evaluates extraction across these structural categories.

We address this gap with the Web Content Extraction Benchmark (WCXB):

\begin{enumerate}
\item \textbf{Scale and diversity.} 2,008 pages from 1,613 domains across seven page types defined by HTML structural differences that affect extraction behavior.
\item \textbf{Page type taxonomy.} A seven-type classification (article, forum, product, collection, listing, documentation, service) grounded in extraction-relevant structural differences rather than semantic content categories.
\item \textbf{Development and test splits.} A 1,497-page development set for system tuning and a 511-page held-out test set with matched page type distribution for generalization validation.
\item \textbf{Comprehensive baselines.} Results from 13 extraction systems --- 11 heuristic and 2 neural --- establishing performance reference points for each page type.
\end{enumerate}

\section{Related Work}

\subsection{Existing Benchmarks}

Table~\ref{tab:benchmarks} summarizes the major web content extraction benchmarks. Most established benchmarks contain fewer than 800 pages, with WebMainBench (7,809 pages, partially released) as the recent exception, and most focus on news articles or mixed content without systematic page type coverage.

\begin{table}[t]
\centering
\caption{Comparison of web content extraction benchmarks.}
\label{tab:benchmarks}
\small
\begin{tabular}{lrllc}
\toprule
\textbf{Benchmark} & \textbf{Pages} & \textbf{Page types} & \textbf{Annotation} & \textbf{Dev/Test} \\
\midrule
CleanEval~\cite{cleaneval} & 797 & mixed (untyped) & manual & no \\
L3S-GN1~\cite{boilerpipe} & 621 & news & manual & no \\
Dragnet~\cite{dragnet} & 143 & articles & manual & no \\
Google-Trends~\cite{boilernet} & 180 & mixed (untyped) & manual & no \\
ScrapingHub~\cite{scrapinghub} & 181 & articles & manual & no \\
WebMainBench~\cite{mineruhtml} & 7,809 & mixed (24 cat.) & manual & partial \\
Bevendorff~\cite{bevendorff} & $\sim$3,985 & inherited mix & combined & no \\
\textbf{WCXB (ours)} & \textbf{2,008} & \textbf{7 labeled types} & \textbf{LLM + human} & \textbf{yes} \\
\bottomrule
\end{tabular}
\end{table}

\textbf{CleanEval}~\cite{cleaneval} was a shared task from 2007. While diverse, its HTML reflects a pre-HTML5 web where semantic markup was rare. \textbf{L3S-GN1}~\cite{boilerpipe}, released with the Boilerpipe system, contains 621 news pages. \textbf{Dragnet}~\cite{dragnet} and \textbf{ScrapingHub}~\cite{scrapinghub} focus exclusively on articles, with 143 and 181 pages respectively. \textbf{Google-Trends-2017}~\cite{boilernet} contains 180 pages used in the BoilerNet evaluation. \textbf{WebMainBench}~\cite{mineruhtml} is the largest existing benchmark at 7,809 pages across 5,434 domains with tag-level HTML annotations and 24 semantic categories. However, it does not provide explicit page type labels for type-stratified evaluation, uses a single evaluation split (no dev/test), and at time of writing the full dataset has not been publicly released --- only a 100-page evaluation subset is available. Our work complements WebMainBench by providing explicit page type labels and a public dev/test split optimized for measuring type-specific extraction quality.

Bevendorff et al.~\cite{bevendorff} made the most comprehensive comparative effort, merging eight datasets and evaluating 14 extractors. Their key finding --- that ``performance is quite genre-dependent'' --- motivates our work. However, their combined dataset does not include new annotations or page type labels, inheriting the article-heavy composition and aging HTML of its sources.

\subsection{Content Extraction Systems}

\textbf{Heuristic systems} dominate production deployments due to speed and reliability. Trafilatura~\cite{trafilatura} combines XPath rules with readability-based fallback. Readability~\cite{readability} (Mozilla) scores DOM nodes by text density. jusText~\cite{justext} classifies text blocks by character-level features. Boilerpipe~\cite{boilerpipe} models extraction as sequence labeling. Resiliparse~\cite{resiliparse} (Webis Group) provides a high-performance C++/Cython HTML parser with main-content heuristics, developed as part of the ChatNoir~\cite{chatnoir} web analytics toolkit; it is the extractor underlying DCLM-Baseline and Li et al.~\cite{li2026beyond} find it particularly effective for table-rich Common Crawl pages. More recent systems include Newspaper4k~\cite{newspaper4k}, Goose3~\cite{goose3}, dom-smoothie~\cite{domsmoothie}, and magic-html~\cite{magichtml}.

\textbf{Neural systems} include BoilerNet~\cite{boilernet}, which applies a sequence labeling neural network to text blocks, and Web2Text~\cite{web2text}, which uses a convolutional neural network over DOM features. More recently, LLM-based systems have emerged: MinerU-HTML~\cite{mineruhtml} (Dripper) fine-tunes Qwen3-0.6B for binary element classification (main vs.\ other), and ReaderLM-v2~\cite{readerlm} (Jina AI) uses a 1.5B parameter model to generate Markdown directly from HTML. These achieve competitive accuracy on articles but require GPU inference, with per-page latencies 36--237$\times$ higher than heuristic approaches.

\subsection{Page Type Classification}

Web page classification has been studied for information retrieval~\cite{qi} and search result presentation~\cite{broder}. Our taxonomy differs from prior genre classification work in that our types are defined by HTML structural properties relevant to extraction --- not by semantic content categories.

\section{Page Type Taxonomy}

We define seven page types based on HTML structural characteristics that require distinct extraction strategies. The taxonomy is driven by extraction behavior: pages are grouped by \emph{how} they need to be extracted, not \emph{what} they are about.

\begin{table}[t]
\centering
\caption{Page type taxonomy with structural characteristics and extraction challenges.}
\label{tab:taxonomy}
\small
\begin{tabular}{p{1.8cm}p{4.8cm}p{5.5cm}}
\toprule
\textbf{Type} & \textbf{Structural characteristics} & \textbf{Extraction challenge} \\
\midrule
Article & Single content container, sequential paragraphs, headings & Boilerplate in sidebars and related article blocks \\
Forum & Multiple user posts, voting/reaction controls, user metadata & Comment-related CSS classes trigger boilerplate removal \\
Product & Structured data (JSON-LD), specifications, pricing, reviews & Content in JSON-LD rather than visible DOM; multiple tabs \\
Collection & Product grids, filter panels, pagination & Filter/navigation panels mixed with content descriptions \\
Listing & Repeated card elements, content index, summaries & Single-node extraction captures one card, not the list \\
Documentation & Code blocks, sidebar navigation, versioned content & TOC and sidebar navigation included in extraction \\
Service & Multi-section layout (hero, features, testimonials, pricing) & Content distributed across 5--15 independent \texttt{<section>} elements \\
\bottomrule
\end{tabular}
\end{table}

This taxonomy emerged empirically from analyzing extraction failures across a web crawl. For example, ``collection'' and ``listing'' both display multiple items, but they differ structurally: collections are product card grids where descriptions appear in JSON-LD or brief overlays, while listings contain substantial text per item (article summaries, course descriptions, review excerpts).

\section{Dataset Construction}

\subsection{Source Selection}

HTML pages were sourced from Common Crawl archives, ChatNoir web search~\cite{chatnoir}, and manual curation to ensure diversity across domains, industries, and geographies. Source URLs were selected to cover all seven page types with emphasis on structural variety within each type --- for example, forum pages span phpBB, Discourse, XenForo, vBulletin, IPS Community, and custom platforms.

The final dataset spans 1,613 unique domains. The median domain contributes 1 page; the 95th percentile is 3 pages. Pages were collected between September 2025 and March 2026, reflecting modern HTML5 markup patterns, responsive design, and contemporary JavaScript frameworks.

\subsection{HTML Collection and Sanitization}

HTML files were downloaded using headless Chrome for JavaScript-rendered pages and direct HTTP requests for server-rendered pages. Sanitization removes tracking scripts, analytics beacons, and advertising tags while preserving content-bearing elements. Character encoding is normalized to UTF-8. Duplicate pages were identified through content hashing and URL deduplication.

Pages where JavaScript rendering is required for content visibility are retained with an SPA (Single Page Application) flag. These pages have empty ground truth content and serve as detection targets --- an extraction system should recognize that no server-side content is available rather than extracting boilerplate.

\subsection{Ground Truth Annotation}

Each ground truth file contains:

\begin{itemize}
\item \textbf{title}: The page's main heading or title
\item \textbf{author}: Author name, when identifiable
\item \textbf{publish\_date}: Publication date, when available
\item \textbf{main\_content}: The complete main content as plain text
\item \textbf{with[]}: Short text snippets (3--8 words) that must appear in a correct extraction
\item \textbf{without[]}: Short text snippets from boilerplate that must not appear in a correct extraction
\end{itemize}

The \texttt{main\_content} field contains plain text with headings on their own lines and paragraphs separated by double newlines. No Markdown or HTML formatting is included --- this ensures evaluation measures content completeness independent of formatting. Content is never truncated. The longest annotation is 49,617 characters.

\subsection{Annotation Methodology}

Ground truth was produced through a multi-stage pipeline:

\textbf{Stage 1: LLM-assisted drafting.} An LLM (Claude) generates initial annotations from the HTML source, extracting title, author, date, and main content. The LLM is instructed to include only content visible in the HTML DOM, excluding navigation, advertisements, and boilerplate.

\textbf{Stage 2: Automated verification.} Scripts verify the generated annotations against the source HTML, checking encoding correctness, JSON schema validity, content bounds, and structural consistency. Files that fail verification are flagged for regeneration or manual correction.

\textbf{Stage 3: Frontier model review.} Each file undergoes four independent review passes by Claude Opus agents, each focusing on different quality dimensions: (1) content completeness, (2) content boundaries, (3) metadata accuracy and snippet quality, (4) internal consistency. Agents compare the ground truth against the source HTML, flag discrepancies, and apply fixes.

\textbf{Stage 4: Snippet and quality verification.} Automated scripts verify that each \texttt{with} snippet appears verbatim in \texttt{main\_content} and no \texttt{without} snippet is present. A 21-point quality scan checks encoding correctness, content bounds, boilerplate contamination, metadata plausibility, and cross-reference integrity. The full scan script is included in the release.

\textbf{Stage 5: Human review.} The human annotator reviews all changes from the preceding stages, adjudicates disagreements between model passes, and performs final quality assurance. Files where the best extraction system achieves low F1 are flagged for manual inspection --- low-scoring files often indicate annotation errors rather than extraction failures.

\subsection{Development and Test Splits}

The dataset is divided into a 1,497-page development set and a 511-page held-out test set. The test set was constructed from a separate pool of pages, reviewed independently, and never used during extraction system development or evaluation-driven tuning. Page type distributions are matched within 4 percentage points between splits (Table~\ref{tab:splits}).

\begin{table}[t]
\centering
\caption{Page type distribution across splits.}
\label{tab:splits}
\small
\begin{tabular}{lrrrr}
\toprule
\textbf{Page type} & \textbf{Dev (N)} & \textbf{Dev (\%)} & \textbf{Test (N)} & \textbf{Test (\%)} \\
\midrule
Article       & 793 & 53.0 & 257 & 50.3 \\
Forum         & 113 & 7.5  & 51  & 10.0 \\
Product       & 119 & 7.9  & 28  & 5.5  \\
Collection    & 117 & 7.8  & 34  & 6.7  \\
Listing       & 99  & 6.6  & 40  & 7.8  \\
Documentation & 91  & 6.1  & 42  & 8.2  \\
Service       & 165 & 11.0 & 59  & 11.5 \\
\midrule
\textbf{Total} & \textbf{1,497} & & \textbf{511} & \\
\bottomrule
\end{tabular}
\end{table}

\section{Dataset Statistics}

\subsection{Content Length}

Content length varies substantially across page types, reflecting genuine structural differences (Table~\ref{tab:contentlen}).

\begin{table}[t]
\centering
\caption{Main content length by page type in characters (development set).}
\label{tab:contentlen}
\small
\begin{tabular}{lrrrr}
\toprule
\textbf{Page type} & \textbf{Min} & \textbf{Median} & \textbf{Mean} & \textbf{Max} \\
\midrule
Article       & 0     & 9,669  & 11,846 & 49,264 \\
Documentation & 607   & 16,385 & 16,632 & 48,790 \\
Service       & 0     & 5,048  & 6,578  & 46,751 \\
Forum         & 0     & 4,806  & 6,024  & 24,571 \\
Listing       & 0     & 2,805  & 6,097  & 37,007 \\
Collection    & 0     & 3,076  & 4,804  & 49,617 \\
Product       & 0     & 2,095  & 2,979  & 13,438 \\
\bottomrule
\end{tabular}
\end{table}

Documentation pages have the highest median content length (16,385 characters), reflecting comprehensive technical references. Product pages have the lowest median (2,095), as many product descriptions are brief. Min~=~0 indicates SPA pages with empty content.

\subsection{Domain Diversity}

The dataset spans 1,613 unique domains across both splits. The development set covers 1,295 domains and the test set covers 472 domains, with 154 domains (9.5\%) appearing in both. Overlapping domains contribute different pages --- typically different page types from the same site.

\section{Baseline Experiments}

\subsection{Evaluation Methodology}

We evaluate extraction quality using word-level F1 score. Extracted text and ground truth are tokenized into word bags (lowercased, splitting on non-alphanumeric characters), and precision, recall, and F1 are computed from word overlap. We chose word-bag F1 over sequence-based metrics (ROUGE-L, edit distance) because content extraction systems vary in whitespace normalization, paragraph ordering, and formatting --- differences that are irrelevant to extraction quality but penalized by order-sensitive metrics. The word-bag approach is consistent with prior extraction benchmarks~\cite{mineruhtml,bevendorff}. Its limitation is insensitivity to word order; we partially address this through \texttt{with}/\texttt{without} snippet evaluation, which tests presence of specific phrases rather than individual words. Per-type 95\% confidence intervals are estimated by non-parametric bootstrap with 1,000 resamples over the per-page F1 scores.

For SPA pages (19 in dev, 4 in test) with empty ground truth, we adopt the convention that a system returning empty text receives F1~=~1.0 (correctly detecting no server-side content) and F1~=~0.0 if boilerplate is extracted. SPA pages have negligible impact on aggregate scores.

\subsection{Systems Evaluated}

We evaluate 13 content extraction systems: 11 heuristic (including one hybrid rule+ML system) and 2 neural. \textbf{Heuristic:} rs-trafilatura~\cite{rstrafilatura}, Resiliparse~\cite{resiliparse}, Trafilatura~\cite{trafilatura}, dom-smoothie~\cite{domsmoothie}, Newspaper4k~\cite{newspaper4k}, magic-html~\cite{magichtml}, jusText~\cite{justext}, Readability~\cite{readability}, BoilerPy3~\cite{boilerpipe} (Python port of Boilerpipe), Goose3~\cite{goose3}, dom-content-extraction~\cite{domcontentextraction}. \textbf{Neural:} MinerU-HTML (Qwen3-0.6B, vLLM on A100)~\cite{mineruhtml}, ReaderLM-v2 (1.5B, vLLM on A100)~\cite{readerlm}.

\textbf{Note on rs-trafilatura:} This system was developed by the first author and iteratively tested against the development set during construction. While the same dev set was used for both GT quality assurance and extraction profile tuning, the held-out test set was never used during development and provides an unbiased comparison. rs-trafilatura uses an XGBoost classifier for page type routing but rule-based heuristics for content extraction itself; we group it with heuristic systems as it does not use neural networks for extraction.

Heuristic systems run on a single machine (AMD Ryzen, 64GB RAM). Neural systems are benchmarked on an NVIDIA A100 GPU with vLLM batched inference.

\subsection{Overall Results}

\begin{table}[t]
\centering
\caption{Extraction systems on the WCXB development set (1,497 pages), sorted by F1.}
\label{tab:overall}
\small
\begin{tabular}{llrrrr}
\toprule
\textbf{System} & \textbf{Type} & \textbf{F1} & \textbf{P} & \textbf{R} & \textbf{ms/p} \\
\midrule
rs-trafilatura      & Rule+ML & \textbf{0.859} & 0.863 & 0.890 & 44 \\
MinerU-HTML (0.6B)  & Neural  & 0.827 & 0.845 & 0.840 & 1,570 \\
Resiliparse         & Rule    & 0.797 & 0.783 & 0.863 & 28 \\
Trafilatura         & Rule    & 0.791 & 0.852 & 0.793 & 97 \\
dom-smoothie        & Rule    & 0.762 & 0.806 & 0.768 & 26 \\
ReaderLM-v2 (1.5B)  & Neural  & 0.741 & 0.741 & 0.790 & 10,410 \\
dom-content-extr.   & Rule    & 0.731 & 0.757 & 0.789 & --- \\
Newspaper4k         & Rule    & 0.720 & 0.838 & 0.683 & 1,825 \\
magic-html          & Rule    & 0.719 & 0.813 & 0.713 & --- \\
jusText             & Rule    & 0.707 & 0.771 & 0.695 & --- \\
BoilerPy3 (article) & Rule    & 0.687 & 0.795 & 0.661 & --- \\
Readability         & Rule    & 0.674 & 0.684 & 0.712 & 785 \\
Goose3              & Rule    & 0.651 & 0.845 & 0.593 & --- \\
\bottomrule
\end{tabular}
\end{table}

rs-trafilatura achieves the highest F1 (0.859) while being among the fastest systems (44ms/page). The gap over Trafilatura (+0.068) comes primarily from non-article page types. MinerU-HTML is a strong competitor at F1~=~0.827, while ReaderLM-v2 (0.741) underperforms several heuristic systems despite its larger model size. Resiliparse (F1~=~0.797), part of the ChatNoir toolkit, edges past Trafilatura overall by trading precision for recall: it captures more content (74.5\% snippet recall vs.\ Trafilatura's 69.5\%) but admits substantially more boilerplate (22.8\% vs.\ 6.6\% \texttt{without} snippet inclusion). Neural model timings are on an NVIDIA A100 with vLLM batched inference.

\textbf{Snippet metrics} provide a complementary view of extraction quality. The \texttt{with} snippet recall measures content completeness (higher is better); the \texttt{without} snippet inclusion rate measures boilerplate contamination (lower is better):

\begin{table}[t]
\centering
\caption{Snippet-based evaluation metrics.}
\label{tab:snippets}
\small
\begin{tabular}{lrr}
\toprule
\textbf{System} & \textbf{With recall} & \textbf{Without incl.} \\
\midrule
rs-trafilatura & \textbf{75.6\%} & 9.6\% \\
Resiliparse    & 74.5\% & 22.8\% \\
Trafilatura    & 69.5\% & 6.6\% \\
Readability    & 67.4\% & 7.5\% \\
dom-smoothie   & 65.6\% & 6.9\% \\
Goose3         & 49.5\% & \textbf{2.4\%} \\
\bottomrule
\end{tabular}
\end{table}

rs-trafilatura achieves the highest content completeness (75.6\% of required snippets found) at the cost of slightly more boilerplate inclusion (9.6\%). Resiliparse reaches comparable content recall (74.5\%) but admits 2--3$\times$ more boilerplate than other top systems, reflecting its design as a fast, broad extractor rather than a precision-oriented one. Goose3's conservative extraction yields the lowest boilerplate contamination (2.4\%) but misses half of required content snippets.

\subsection{Per-Page-Type Results}

The per-type breakdown reveals the key finding motivating this benchmark (Table~\ref{tab:pertype}).

\begin{table}[t]
\centering
\caption{F1 by page type for top extraction systems (development set). Best value per row in bold.}
\label{tab:pertype}
\small
\begin{tabular}{lrrrrrrr}
\toprule
\textbf{Page type} & \textbf{N} & \textbf{rs-traf} & \textbf{MinerU} & \textbf{Resil.} & \textbf{RdrLM} & \textbf{Trafil.} & \textbf{d-smo.} \\
\midrule
Article       & 793 & \textbf{0.932} & 0.928 & 0.871 & 0.880 & 0.924 & 0.903 \\
Documentation & 91  & \textbf{0.931} & 0.838 & 0.883 & 0.776 & 0.888 & 0.868 \\
Service       & 165 & \textbf{0.843} & 0.824 & 0.815 & 0.708 & 0.751 & 0.721 \\
Forum         & 113 & 0.792 & \textbf{0.794} & 0.758 & 0.589 & 0.575 & 0.520 \\
Collection    & 117 & \textbf{0.713} & 0.506 & 0.586 & 0.415 & 0.518 & 0.480 \\
Listing       & 99  & 0.704 & \textbf{0.710} & 0.620 & 0.578 & 0.550 & 0.544 \\
Product       & 119 & \textbf{0.670} & 0.619 & 0.608 & 0.464 & 0.562 & 0.489 \\
\bottomrule
\end{tabular}
\end{table}

Bootstrap 95\% confidence intervals for rs-trafilatura range from $\pm$0.01 on articles (N=793) to $\pm$0.06 on forums (N=113) and listings (N=99), reflecting the smaller sample sizes for non-article types. Per-type comparisons on categories with fewer than 120 pages should be interpreted with this uncertainty in mind.

On articles, all systems in Table~\ref{tab:pertype} converge within a narrow band (F1~=~0.871--0.932). The picture changes dramatically on other page types:

\begin{itemize}
\item \textbf{Forums}: A 27.4-point spread (0.520--0.794). Systems that treat comment-related CSS classes as boilerplate lose the primary content of forum threads.
\item \textbf{Collections}: A 29.8-point spread (0.415--0.713). Product grid pages confuse extractors with interleaved filter panels and navigation.
\item \textbf{Products}: A 20.6-point spread (0.464--0.670). Content encoded in JSON-LD structured data is invisible to DOM-only extractors.
\item \textbf{Service pages}: A 13.5-point spread (0.708--0.843). Content distributed across multiple \texttt{<section>} elements defeats single-node extraction.
\end{itemize}

These gaps are invisible to article-only benchmarks.

\subsection{Generalization: Held-Out Test Set}

To validate that results generalize beyond the development set, we evaluate the top systems on the 511-page held-out test set (Table~\ref{tab:heldout}).

\begin{table}[t]
\centering
\caption{Generalization results (held-out test set, 511 pages).}
\label{tab:heldout}
\small
\begin{tabular}{lrrr}
\toprule
\textbf{System} & \textbf{Dev F1} & \textbf{Test F1} & \textbf{$\Delta$} \\
\midrule
rs-trafilatura & 0.859 & \textbf{0.903} & +0.044 \\
Resiliparse    & 0.797 & 0.817 & +0.020 \\
Trafilatura    & 0.791 & 0.841 & +0.050 \\
dom-smoothie   & 0.762 & 0.817 & +0.055 \\
Newspaper4k    & 0.720 & 0.762 & +0.042 \\
Readability    & 0.674 & 0.736 & +0.062 \\
Goose3         & 0.651 & 0.694 & +0.043 \\
\bottomrule
\end{tabular}
\end{table}

All systems score higher on the test set than the development set, with the performance \emph{ranking} largely preserved. The shifts span +0.020 to +0.062, reflecting two factors: (1) the test set's GT annotations were produced by a more refined pipeline, and (2) the test set has a slightly higher proportion of articles (50.3\% vs.\ 53.0\% in dev), which are the easiest page type. One re-ranking is notable: on the held-out set, Trafilatura (0.841) and dom-smoothie (0.817) overtake Resiliparse (0.817), reversing their dev-set order --- Resiliparse benefits least from the cleaner test annotations (smallest $\Delta$~=~+0.020), consistent with its higher boilerplate inclusion rate. The leading system is stable across splits: rs-trafilatura leads the second-place system by 6.2 points on dev (vs.\ Resiliparse) and by 6.2 points on test (vs.\ Trafilatura), confirming that relative comparisons among the top systems generalize.

The neural systems (MinerU-HTML, ReaderLM-v2) were evaluated on the development set only due to GPU compute constraints; their generalization to the held-out set is left for future work. Given the article-favoring composition of the held-out set, we expect both systems to show similar upward shifts on articles with narrower gains on the structured page types where they already underperform heuristics.

\subsection{Cross-Benchmark Validation}

To verify consistency with existing evaluations, we evaluate on the ScrapingHub article extraction benchmark~\cite{scrapinghub} (181 pages, articles only), using its native 4-gram shingle F1 metric (Table~\ref{tab:crossbench}).

\begin{table}[t]
\centering
\caption{Cross-benchmark results on ScrapingHub (181 articles).}
\label{tab:crossbench}
\small
\begin{tabular}{lrr}
\toprule
\textbf{System} & \textbf{WCXB dev F1} & \textbf{ScrapingHub F1} \\
\midrule
rs-trafilatura & 0.859 & \textbf{0.967} \\
Trafilatura    & 0.791 & 0.958 \\
\bottomrule
\end{tabular}
\end{table}

Both systems score substantially higher on ScrapingHub, as expected for an article-only benchmark. On WCXB's article subset alone, rs-trafilatura achieves F1~=~0.932 --- close to its 0.967 on ScrapingHub. The gap between WCXB overall F1 (0.859) and article-only F1 (0.932) quantifies the impact of page type diversity on benchmark scores: including non-article pages reveals a 7.3-point performance gap that article-only benchmarks cannot detect.

\section{Analysis}

\subsection{Article Extraction Is Largely Solved}

On articles, every system in Table~\ref{tab:pertype} achieves F1~$\geq$~0.871, and four of the six exceed F1~=~0.90 (rs-trafilatura 0.932, MinerU-HTML 0.928, Trafilatura 0.924, dom-smoothie 0.903). The spread between the best heuristic (rs-trafilatura) and the best neural system (MinerU-HTML) is only 0.4 F1 points. This convergence has an important implication: \textbf{article-only benchmarks have diminishing discriminative power.} The performance ceiling on articles has been reached, and future progress in extraction quality must come from other page types.

\subsection{Structured Page Types Remain Open}

In contrast, no system exceeds F1~=~0.80 on forums, 0.72 on collections, or 0.68 on products. These are not marginal categories --- in our dataset, 47\% of pages are non-articles. The failures are structural rather than statistical: forum extraction requires recognizing comment elements as content, product extraction requires JSON-LD fallback, and service page extraction requires multi-section merging. These failure modes suggest that improving extraction on structured page types requires \emph{architectural changes} rather than parameter tuning.

\subsection{Neural Systems Do Not Solve the Diversity Problem}

A common assumption is that neural approaches should generalize better across page types. Our results challenge this: MinerU-HTML trails the best heuristic on 6 of 7 page types, and ReaderLM-v2 underperforms three heuristic systems overall despite being 2.5$\times$ larger. On collections, both neural systems score below all top heuristics (ReaderLM-v2: 0.415, MinerU-HTML: 0.506, vs.\ rs-trafilatura: 0.713). Neural systems inherit the same article bias as heuristic systems --- they were trained predominantly on article-like content. Larger model size (1.5B vs.\ 0.6B) does not help: ReaderLM-v2 scores lower than MinerU-HTML on every page type.

\subsection{Speed Considerations}

For production deployment, extraction speed matters. Heuristic systems process pages in 26--1,825ms on commodity hardware, with most under 100ms. Neural systems require 1,570--10,410ms per page on an A100 GPU. At web-crawl scale, this translates to order-of-magnitude cost differences.

\section{Limitations}

\textbf{Language coverage.} The dataset is predominantly English. Extraction systems may perform differently on non-English content. Multilingual expansion is planned for future versions.

\textbf{Static HTML snapshots.} All pages are server-rendered HTML snapshots. JavaScript-heavy SPAs are represented only by flagged empty-content entries.

\textbf{LLM-assisted annotation pipeline.} Initial drafts were generated by an LLM (Claude) and reviewed through four passes by Claude Opus agents, with automated verification scripts running before and after the model review stage. A single human annotator reviewed all changes, adjudicated disagreements, and performed final quality assurance. We cannot report formal inter-annotator agreement metrics. The LLM-driven stages may introduce systematic biases; we mitigate this through multi-pass structure, two rounds of automated verification, and adversarial review against extraction results, but subjectivity in content boundary decisions remains unquantified.

\textbf{Page type balance.} Articles comprise 53\% of the dataset. Smaller categories (documentation: 6.1\%, listing: 6.6\%) have fewer examples. Per-type results for these categories should be interpreted with sample size in mind.

\textbf{Temporal snapshot.} HTML collected between September 2025 and March 2026 reflects current web design patterns but may not represent future trends.

\section{Ethics and Data Access}

The dataset is released under CC-BY-4.0. The release includes HTML source files, JSON ground truth annotations, page type labels, development/test split assignments, and baseline extraction results. Ground truth annotations contain only content visible on publicly accessible web pages. No personally identifiable information is included beyond what appears in the original public web content.

The dataset is available at:
\begin{itemize}
\item \textbf{GitHub}: \url{https://github.com/Murrough-Foley/web-content-extraction-benchmark}
\item \textbf{Zenodo}: \url{https://doi.org/10.5281/zenodo.19316874}
\item \textbf{HuggingFace}: \url{https://huggingface.co/datasets/murrough-foley/web-content-extraction-benchmark}
\end{itemize}

\section{Conclusion}

We present WCXB, a 2,008-page benchmark for web content extraction that addresses the article-only blind spot of existing datasets. By splitting the data into a 1,497-page development set and a 511-page held-out test set and by labeling every page with one of seven structural types, the benchmark supports both system development and generalization testing, and exposes per-type performance that aggregate scores obscure. Our baseline evaluation of 13 extraction systems shows that article extraction is effectively saturated (F1~$\geq$~0.87 for every system in Table~\ref{tab:pertype}) while forums, collections, and products remain open problems (F1 gaps of 20--30 points between systems). Neural approaches do not close these gaps: both MinerU-HTML and ReaderLM-v2 inherit the article bias of their training data. Held-out results preserve the development-set ranking with a uniform upward shift, supporting the generalization of relative comparisons. We release the dataset, evaluation toolkit, and baseline results under CC-BY-4.0 to support reproducible research on type-aware extraction.

\bibliographystyle{plain}

\end{document}